# A Biologically Inspired Separable Learning Vision Model for Real-time Traffic Object Perception in Dark

Hulin Li[1, 2, *], Qiliang Ren[1, *], Jun Li[3], Hanbing Wei[3], Zheng Liu[2], Linfang Fan[1]

1 School of Traffic and Transportation, Chongqing Jiaotong University, Chongqing 400074, China;
2 School of Engineering, University of British Columbia, Okanagan, Kelowna BC V1V 1V7, Canada;
3 School of Mechatronics and Vehicle Engineering, Chongqing Jiaotong University, Chongqing 400074, China

**alan@mails.cqjtu.edu.cn (H.L.);** qlren@cqjtu.edu.cn (Q.R.); cqleejun@cqjtu.edu.cn (J.L.);
**zheng.liu@ubc.ca (Z.L.);** hbwei@cqjtu.edu.cn (H.W.); linfangfan@cqjtu.edu.cn (L.F)

**Abstract:**

Fast and accurate object perception in low-light traffic scenes has attracted increasing attention. However, due to severe illumination degradation and the lack of reliable visual cues, existing perception models and methods struggle to quickly adapt to and accurately predict in low-light environments. Moreover, there is the absence of available large-scale benchmark specifically focused on low-light traffic scenes. To bridge this gap, we introduce a physically grounded illumination degradation method tailored to real-world low-light settings and construct Dark-traffic, the largest densely annotated dataset to date for low-light traffic scenes, supporting object detection, instance segmentation, and optical flow estimation. We further propose the Separable Learning Vision Model (SLVM), a biologically inspired framework designed to enhance perception under adverse lighting. SLVM integrates four key components: a light-adaptive pupillary mechanism for illumination-sensitive feature extraction, a feature-level separable learning strategy for efficient representation, task-specific decoupled branches for multi-task separable learning, and a spatial misalignment-aware fusion module for precise multi-feature alignment. Extensive experiments demonstrate that SLVM achieves state-of-the-art performance with reduced computational overhead. Notably, it outperforms RT-DETR by 11.2 percentage points in detection, YOLOv12 by 6.1 percentage points in instance segmentation, and reduces endpoint error (EPE) of baseline by 12.37% on Dark-traffic. On the LIS benchmark, the end-to-end trained SLVM surpasses Swin Transformer+EnlightenGAN and ConvNeXt-T+EnlightenGAN by an average of 11 percentage points across key metrics, and exceeds Mask RCNN (with light enhancement) by 3.1 percentage points. The Dark-traffic dataset and complete code is released at https://github.com/alanli1997/slvm. **The journal version has been published by Elsevier in *Expert Systems with Applications*. Please cite the journal version, DOI:** https://doi.org/10.1016/j.eswa.2025.129529

## 1. Introduction

Real-time perception in low-light transportation environments is essential for a wide range of engineering applications, including advanced driver assistance systems, visual navigation, traffic surveillance, and energy-efficient smart sensing [50]. While visual models under well-lit conditions have achieved saturated performance, low-light scenes still pose substantial challenges [1]. These difficulties are primarily due to three key factors. First, photon-starvation-induced boundary ambiguity: the scarcity of photons in low-light environments leads to highly entropic image signals and blurred object-background boundaries, making accurate foreground extraction difficult. Second, decoupled enhancement pipelines: most low-light image enhancement techniques are computationally expensive and task-inconsistent, preventing seamless integration with real-time perception systems. Third, limited real-world annotations: existing datasets lack fine-grained, pixel-level labels that reflect the complex illumination characteristics of real-world low-light traffic scenes.

These issues underscore the need for a paradigm shift in both model architecture and dataset development.

Through extensive investigation, we identify a significant disparity between the visual mechanisms and learning strategies employed by existing general real-time object perception models and those observed in intelligent biological systems [2, 3]. In the domain of visible-light vision, many intelligent species—such as dogs and primates—exhibit a remarkable ability to dynamically adjust their visual processing in response to variations in ambient illumination. In contrast, modern object perception models lack such adaptive capabilities, operating under static inference pipelines regardless of environmental lighting. Moreover, most state-of-the-art perception models adopt a feature pyramid architecture [5–7], where the stage-scale features are generated in a one-pass fashion and shared across different visual tasks. This monolithic, task-agnostic feature utilization strategy stands in stark contrast to two fundamental principles widely observed in biological learning systems. First, intelligent agents tend to decompose complex tasks into simpler sub-tasks during the learning process [8]. Second, feature representations in biological vision are highly adaptive, modulating based on the specific demands of the task [9].

Motivated by these observations, we propose the SLVM, a biologically inspired framework designed to tackle the challenges of low-light traffic perception through three key components. The first, light-adaptive pupillary perception mechanism (LAPM), emulates the rapid pupillary response in biological systems to improve photoreceptive encoding under varying illumination. The second, feature-level separable learning strategy (FSLConv), decouples texture features at the same scale, enabling more efficient learning and greater representation diversity. The third, task-specific adaptive learning strategy, uses independent and specific branches to separately learn illumination-aware and semantic features, enhancing adaptability across perception tasks.

To support this framework, we construct Dark-traffic, the largest publicly available dataset for low-light transportation perception to date. It is generated using an empirical illumination-degradation pipeline derived from real-world low-light distributions and contains over 10,000 images with 100,000 annotated instances across object detection, instance segmentation, and optical flow tasks.

In addition, we introduce a feature-refined soft nearest neighbor interpolation technique (SNI-r) that addresses misalignment in multi-scale fusion, further improving perceptual robustness under degraded lighting. Our framework is evaluated on both Dark-traffic and existing low-light benchmarks, demonstrating state-of-the-art performance across static and dynamic tasks.

Our main contributions are summarized as follows:

(1) We release a new large-scale dataset for object detection, instance segmentation, and optical flow in low-light traffic scenes to support future research.

(2) We introduce a biologically inspired perception framework, including light-adaptive pupillary perception mechanism, feature-level separable learning convolution, and task-decomposed strategy of semantic and photoreceptive.

(3) We propose SNI-r, a feature-level interpolation method that mitigates misalignment in multi-scale fusion.

(4) We conduct extensive experiments across multiple low-light benchmarks, demonstrating superior performance and robustness. Both the dataset and code will be released to foster further development in this field.

The remainder of this paper is organized as follows: Section 2 reviews related work. Section 3 presents the Dark-traffic dataset and the components of SLVM. Section 4 provides experimental results and analysis. Section 5 discusses broader implications and limitations, and Section 6 concludes all the work.

## 2. Related work

### 2.1 Low-light object perception datasets

Despite growing interest in robust perception under poor illumination, public datasets for low-light traffic scenes remain scarce—especially for pixel-level tasks like instance segmentation and optical flow. Existing datasets such as ExDark [10], LIS [1], and recent work by Zhang et al. [11], Chen et al. [12], and Zheng et al. [13] are limited in scale, coverage, or annotation quality. Among them, only LIS provides instance-level labels and traffic-specific categories. However, none comprehensively support dense annotations under real low-light traffic conditions. Datasets designed for image enhancement [16, 17] are also of limited value here, as they target human-perceivable restoration rather than task-relevant representation learning. To bridge this gap, we introduce Dark-traffic, the first large-scale dataset offering dense annotations for detection, instance segmentation, and motion estimation in real-world low-light traffic environments.

## 2.2 Low-light object perception methods

Real-world low-light perception is inherently challenging due to photon scarcity, visual noise, and severe texture degradation. Traditional approaches have predominantly focused on low-light image enhancement [22, 49], utilizing methods such as Retinex decomposition [14, 15], illumination-aware tone mapping, and generative adversarial networks (e.g., EnlightenGAN [22]). While these techniques are effective in improving perceptual quality, they are often computationally intensive and task-agnostic, making them unsuitable for real-time, end-to-end visual recognition systems. Recent attempts to integrate enhancement with downstream perception tasks (e.g., detection and segmentation [11, 12]) show promise, but still suffer from several limitations: most rely on multi-stage pipelines, require manual supervision for enhancement, or fail to generalize across tasks. These constraints hinder their deployment in unified, real-time, and task-aware low-light perception systems.

## 2.3 Object perception model architectures

Network architecture design generally follows two paths: manual design and neural architecture search (NAS). NAS leverages reinforcement learning to generate hardware-specific structures, but currently suffers from high computational cost, unstable generalization, poor interpretability, and limited scalability [18]. In contrast, manually designed architectures embed domain knowledge and often generalize better, offering reusable design insights. However, some widely accepted design heuristics now require re-evaluation as models evolve [21, 54]. For example, guidelines from Inceptionv2 [19] and ShuffleNetv2 [20]—such as balancing depth and width, or reducing fragmentation—have shown limited validity in later work like VanillaNet [52] and CSPNet [30], which proposed more effective alternatives. The proposed SLVM framework builds on biological vision principles, incorporating interpretable modules—FSLConv (theoretically motivated), LAPM (adaptive luminance modulation), and SNI-r (fine-grained feature alignment). Details are elaborated in Section 3.

A summary of related work and the specific gaps this study aims to fill is presented in Table 1.

Table 1. Comparison of existing work and identified research gaps addressed by this work

| Category | Representative Work | Key Contributions | Limitations | This Work Addresses |
| --- | --- | --- | --- | --- |
| Low-light datasets | LIS [1], ExDark [10], Zhang et al. [11], etc. | Provide datasets under low-light conditions; LIS offers instance-level traffic annotations | Datasets are small, lack dense annotations for segmentation and motion; not tailored to real traffic scenarios | Introduces Dark-traffic, a large-scale dataset with dense annotations for detection, segmentation, and flow |
| Enhancement models | Retinex [14,15], EnlightenGAN [22], REDI [49] | Improve perceptual quality via image enhancement before perception tasks | Multi-stage, task-agnostic, computationally heavy; poor generalization in real-time unified perception | Focuses on task-aware, end-to-end low-light perception without explicit enhancement |
| Integrated perception models | Zhang et al. [11], Chen et al. [12], Zheng et al. [13] | Fuse enhancement and object perception in unified models | Require manual supervision; limited scalability; lack biological interpretability; low task generalization | Proposes SLVM, a biologically inspired, interpretable framework with modular, efficient design |
| Architecture design | NAS [18], Inceptionv2 [19], ShuffleNetv2 [20], CSPNet [30], VanillaNet [52] | Explore efficient designs via experience, heuristics or search | Some experience outdated; NAS costly and unstable; limited task-specific adaptability | Designs FSLConv, LAPM, and SNI-r based on interpretable biological principles |

## 3. Methods

This section presents the construction details of the Dark-traffic dataset and the proposed SLVM framework, including: (1) a physically grounded illumination degradation model based on real low-light scenarios, (2) the biologically inspired LAPM mechanism for light-sensitive feature extraction, (3) the FSLConv module for feature-level decoupled learning inspired by animal learning behaviors, (4) a task-specific branch design motivated by perceptual specialization in biological systems, and (5) the SNI-r module, which addresses misalignment in multi-level feature fusion under spatial and task discrepancies.

## 3.1 Illumination degradation transfer and construction of the Dark-traffic dataset

Object perception in low-light transportation scenes presents two major challenges. First, existing datasets (e.g., LIS [1]) are limited in scale and diversity, restricting the evaluation of model generalization under real-world conditions. Second, collecting pixel-level annotations for instance segmentation and optical flow is both time-consuming and costly. While generative adversarial networks (GANs) and diffusion models have been explored to synthesize low-light images [22, 23], these approaches are primarily designed for visual enhancement, which fundamentally differs from perception-oriented tasks. Enhancement methods typically focus on optimizing pixel-wise reconstruction for improved human visual quality, whereas perception tasks prioritize feature-level readability and task-specific performance.

To address this gap, we propose a feature-preserving illumination degradation method specifically tailored for visual perception tasks. Compared to generative approaches, our method offers three key advantages: Task-Oriented Design-It breaks away from pixel-level enhancement constraints and focuses on aligning degraded images with the feature space requirements of perception models; Physical Interpretability-It models illumination degradation based on statistical distributions derived from real-world low-light traffic data, ensuring alignment with practical operational scenarios; Efficient Scalability-It supports domain adaptation for arbitrary instance segmentation datasets without requiring additional manual labeling, thus enabling large-scale low-light dataset generation at near-zero marginal cost.

Unlike enhancement-driven methods, our illumination degradation process does not prioritize visual fidelity from a human-centric perspective, and may result in perceptual color distortions. However, this trade-off is intentional and acceptable in the context of feature-centric vision models. To ensure the realism of degraded samples, we construct paired source-target datasets using real-world low-light image samples to supervise the degradation transfer. Based on this, we establish a new large-scale dataset—Dark-traffic—designed for object detection, instance segmentation, and optical flow estimation in low-light transportation scenarios. The complete data construction pipeline is illustrated in Figure. 1.

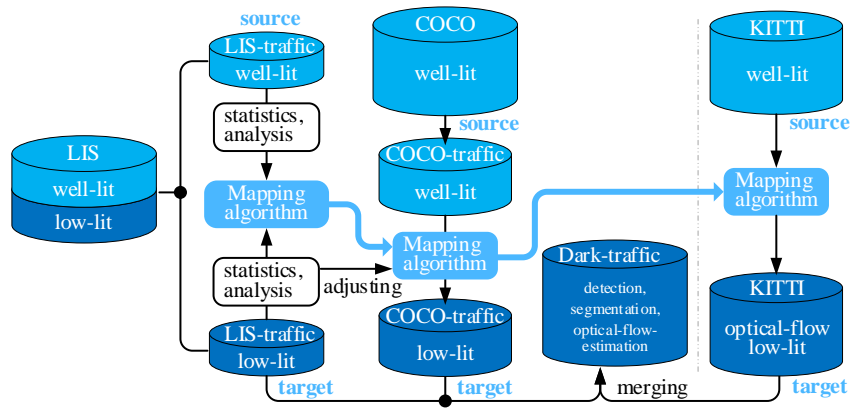

**Figure 1.** Dark-traffic dataset synthesis pipeline for low-light traffic scene perception via illumination style transfer.

We first curated traffic-relevant samples from the real-world LIS dataset [1], manually adding annotations for underrepresented but critical classes such as person and traffic light. To analyze illumination-dependent characteristics, we conducted RGB-channel-wise feature statistics on both well-lit and low-lit images, obtaining their empirical distributions in the RGB color space (Figure. 2).

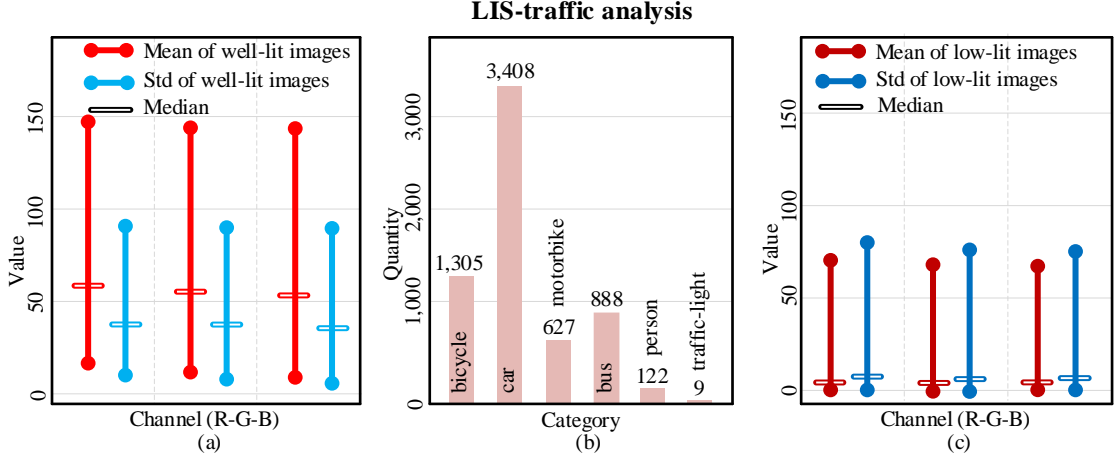

**Figure 2.** LIS-traffic analysis: (a) RGB value distribution of well-lit images. (b) Class-wise annotation distribution for the six traffic-related categories (unbalance). (c) RGB value distribution of low-lit images.

These distributions form the foundation for constructing a feature-aligned illumination degradation mapping that transfers images from well-lit (source domain) to low-lit (target domain) conditions. Specifically, we adopt a statistically grounded transformation framework comprising the following steps:

(I). Truncated normal sampling of target statistics

For each color channel $C \in \{R, G, B\}$ - representing the red (R), green (G), and blue (B) components of the standard additive RGB color space - we estimate the target mean $\mu_t^C$ and standard deviation $\sigma_t^C$ by sampling from truncated normal distributions fitted to the empirical statistics of the LIS-traffic (low-light) dataset:

$$\mu_t^C \sim \mathcal{TN}(a_\mu^C, b_\mu^C, \mu_{med}^C, \sigma_\mu^C), \sigma_t^C \sim \mathcal{TN}(a_\sigma^C, b_\sigma^C, \sigma_{med}^C, \sigma_\sigma^C)$$

where $\mathcal{TN}$ denotes the truncated normal distribution; $\mu_{med}^C$ and $\sigma_\mu^C$ are the median and standard deviation of the observed means for channel $C$; $\sigma_{med}^C$ and $\sigma_\sigma^C$ are the median and standard deviation of the observed standard deviations; $a_\mu^C$ and $b_\mu^C$ are the lower and upper truncation boundaries for the mean sampling, calculated as:

$$a_\mu^C = \frac{\mu_{min}^C - \mu_{med}^C}{\sigma_\mu^C}, b_\mu^C = \frac{\mu_{max}^C - \mu_{med}^C}{\sigma_\mu^C}$$

Similarly, $a_\sigma^C$ and $b_\sigma^C$ are derived for the standard deviation sampling. $\mu_{min}^C$ and $\mu_{max}^C$ denote the minimum and maximum observed per-image mean intensities for channel $C$ respectively.

(II). Adaptive channel-wise linear transformation

For each input image channel $I^C \in \mathbb{R}^{H \times W}$, where $H$ and $W$ denote image height and width in pixels, we apply a linear transformation that aligns both contrast and brightness to the sampled target statistics as Equation (1):

$$I_{adj}^C = \text{clip}_{[0,255]}(\frac{\sigma_t^C}{\sigma_o^C}(I^C - \mu_o^C) + \mu_t^C) \quad (1)$$

where $I_{adj}^C$ is the adjusted image channel after illumination degradation; $\mu_o^C$ and $\sigma_o^C$ are the mean and standard deviation of the original channel $C$, computed as:

$$\mu_o^C = \frac{1}{HW}\sum_{i=1}^{H}\sum_{j=1}^{W} I^C(i,j), \sigma_o^C = \sqrt{\frac{1}{HW}\sum_{i=1}^{H}\sum_{j=1}^{W}(I^C(i,j) - \mu_o^C)^2}$$

The $\text{clip}_{[0,255]}(x)$ ensures that the transformed pixel intensities remain within valid image range [0, 255].

(III). Color consistency preservation

To mitigate perceptual color distortions introduced by independent channel-wise adjustments, we define color ratio matrices for both original and adjusted images:

$$R_o(i,j,c) = \frac{I_o^c(i,j)}{\sum_{k \in \{R,G,B\}} I_o^k(i,j) + \epsilon}, R_{adj}(i,j,c) = \frac{I_{adj}^c(i,j)}{\sum_{k \in \{R,G,B\}} I_{adj}^k(i,j) + \epsilon}$$

where $R_o(i,j,c)$ and $R_{adj}(i,j,c)$ denote the normalized color proportions at pixel $(i,j)$ for channel c in the original and adjusted images, respectively; $I_o^c(i,j)$ is the original image channel at pixel $(i,j)$; $\epsilon$ is a small constant ($10^{-8}$) to prevent division by zero.

Based on these ratios, we compute a binary correction mask $M(i, j)$ that identifies pixels requiring color adjustment as Equation (2):

$$M(i,j) = \begin{cases} 1, \text{ if } max_c|R_o(i,j,c) - R_o(i,j,c)| > \tau \\ 0, \text{ otherwise} \end{cases} \quad (2)$$

where $\tau$ is the color consistency threshold, empirically set to 0.5.

The final pixel value is computed as Equation (3):

$$I_f^C(i,j) = \begin{cases} R_o(i,j,c)\sum_k I_{adj}^k(i,j), \text{ if } M(i,j) = 1 \\ I_{adj}^C(i,j), \text{ otherwise} \end{cases} \quad (3)$$

where $I_f^C(i,j)$ denotes the final corrected value for channel $C$ at pixel $(i, j)$.

We apply this transformation pipeline to a well-lit subset of COCO [36], which we denote as COCO-traffic (well-lit), curated by selecting only traffic-participant categories. This yields a low-light counterpart, COCO-traffic (low-lit), whose color distribution closely resembles that of real-world low-light scenes (see Figure. 3).

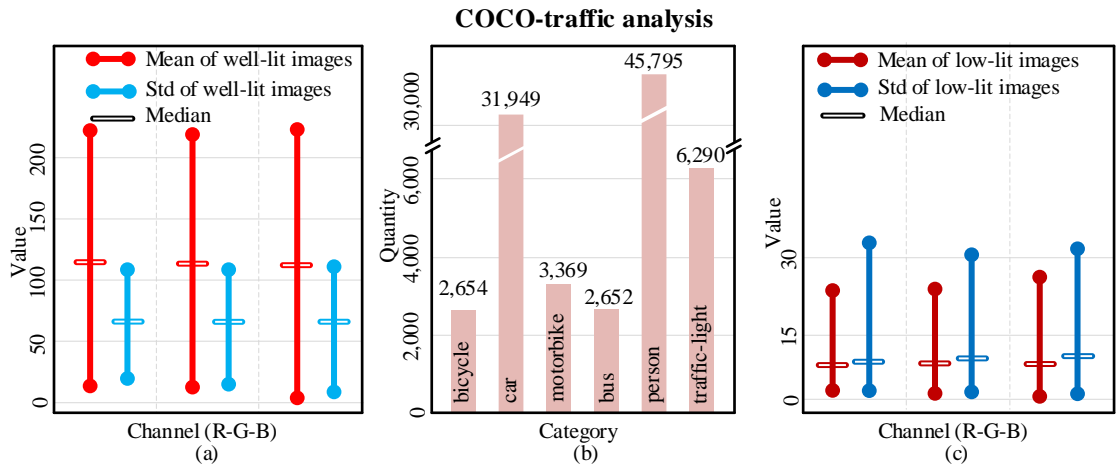

**Figure 3.** COCO-traffic analysis: (a) RGB value distribution of well-lit images. (b) Class-wise annotation distribution for the six traffic-related categories (balance). (c) RGB value distribution of low-lit images.

Finally, we merge the transformed COCO-traffic (low-lit) with LIS-traffic (low-lit) to construct the large-scale Dark-traffic dataset, consisting of 10,425 low-light images and 99,014 instance-level pixel annotations. The category distribution and image-wise statistics are illustrated in Figure. 4(a) and 4(b), respectively, demonstrating strong alignment with the empirical distribution of real low-light images in Figure. 2(c).

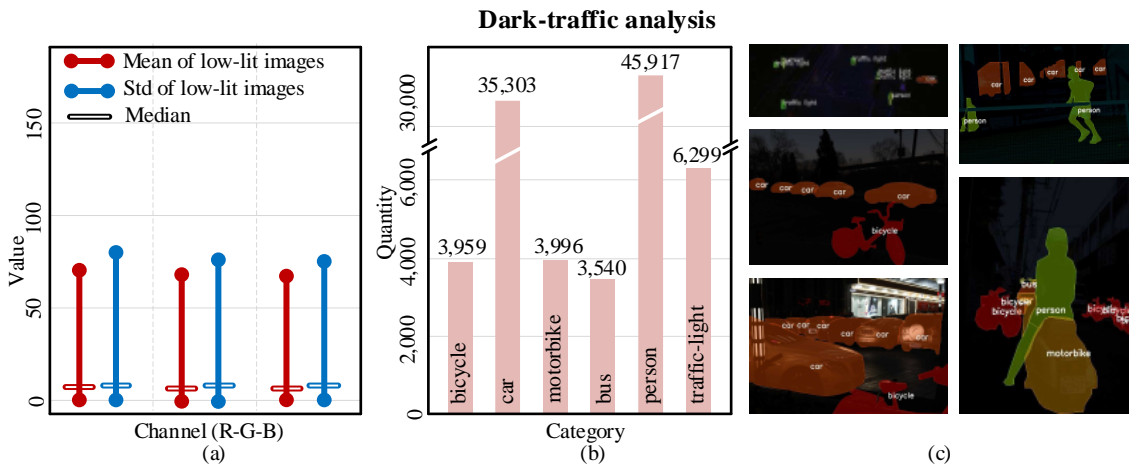

**Figure 4.** Dark-traffic analysis: (a) RGB value distribution of low-lit images. (b) Class-wise annotation distribution for the six traffic-related categories (balance). (c) Example of low-light traffic instance segmentation samples of Dark-traffic.

It is worth noting that many previous work regard synthetic noise injection as a key step in generating low-light images. However, we argue that this practice should be approached with caution for the following reasons. First, due to complex physical phenomena such as photon shot noise, scattering, and surface reflection under insufficient illumination, the noise patterns captured by optical sensors in real low-light conditions are inherently diverse and poorly defined. This makes it extremely difficult to formulate a physically meaningful and universally valid definition of "noise" in such settings. When the very definition is ambiguous, injecting artificial noise amounts to data contamination, potentially degrading downstream model performance [24]. Second, the effectiveness of smoothing filters against classical Gaussian noise is already well understood, making it unnecessary and even redundant to artificially create a problem only to resolve it with conventional methods. Thus, we discourage adding synthetic noise to collected image datasets. The Dark-traffic dataset does not involve any artificial noise injection. However, it includes a subset of short-exposure raw dark images that naturally contain sensor-induced noise. In experiments, these images do not exhibit any detrimental impact on model training or performance. All datasets, preprocessing code, and domain transfer scripts will be publicly available to facilitate reproducibility and further research.

### 3.2 Biological inspired light-adaptive pupillary perception mechanism

A key observation derived from the statistical analysis in Section 3.1 is that the RGB channel values of images captured under low-light conditions are significantly lower than those captured under well-lit conditions. From a biological perspective, this reflects a substantial reduction in the number of photons received by the visual system. Conventional approaches in computer vision typically address this issue by enhancing the illumination using Retinex-like models [14], or by training supervised or unsupervised enhancement networks using paired low-light and normal-light images [22, 23]. These methods aim to restore images toward human visual perception and fidelity.

However, such approaches are often suboptimal for real-time perception tasks in low-light transportation scenes. On the one hand, detection and segmentation models rely more on texture-level features than fine image details. On the other hand, image enhancement algorithms that focus on restoring visual fidelity often introduce significant computational overhead, compromising the inference speed and efficiency of downstream models.

In contrast, biological vision systems handle low-light conditions in a far more efficient manner—primarily through pupillary dilation, which regulates the light flux entering the eye. For instance, the human pupil diameter ranges from ~ 1mm to ~ 9mm, while cats can adjust between ~ 0.5mm and ~ 12mm. Notably, minor color distortions do not significantly impair object perception in biological systems [25]. Inspired by this observation, we propose the LAPM, which simulates pupillary dilation using an amplification factor to compensate for illumination loss. In addition, we design a characterized photosensitive mask to extract texture-aware features in low-light conditions at an extremely low computational cost. The complete computational pipeline is illustrated in Figure. 5(g), with detailed mathematical formulations provided in the supplementary material.

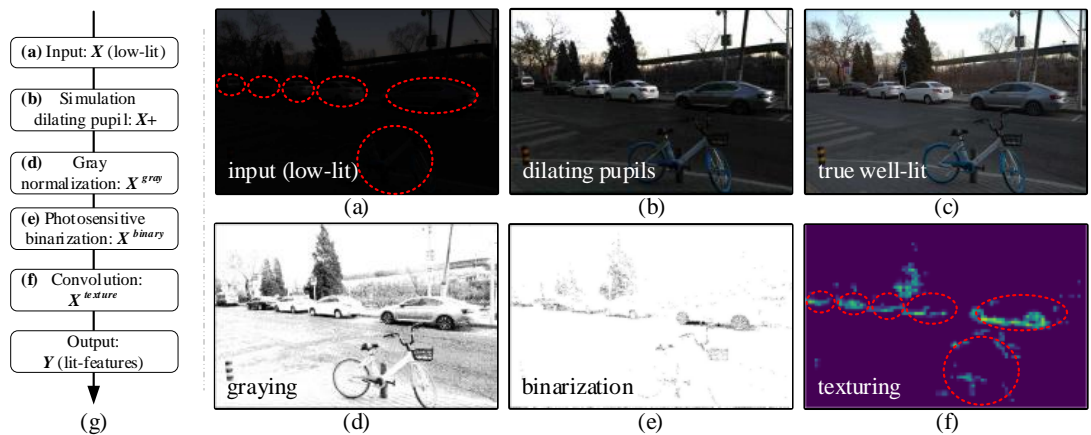

**Figure 5.** Computational pipeline (g) and visualization (a-e) of the LAPM: (a) Input low-light image sample. (b) Simulated pupillary dilation-based luminance-enhanced image. (c) Ground-truth well-lit counterpart. (d) Grayscale version of the (b). (e) Photoreceptive binary mask generated from the (d). (f) Photoreceptive feature mask extracted from the (e). (g) Full computational pipeline of the LAPM.

Specifically, after input normalization and before feeding images into the network, we amplify the pixel values to simulate the optical gain introduced by dilation. While this may introduce perceptual distortions in the image (see Figure. 5(b), (c)), it does not significantly alter the underlying texture structures, which are essential for detection and segmentation. The compensated image is then partially converted to grayscale (Figure. 5(d)) to preserve structural cues under luminance enhancement. A photon threshold is applied to binarize the grayscale map into a photosensitive mask (Figure. 5(e)), which is further processed using a lightweight convolutional kernel to extract high-level light-compensated texture features (Figure. 5(f)).

This module introduces only four trainable parameters and incurs a minimal computational overhead of 0.002184 GFLOPs (for 640×640 resolution input). Yet, it effectively enhances foreground texture representations. For example, Figure. 5(f) clearly demonstrates how the local textures of five cars and one bicycle are successfully separated from the background, despite the challenging lighting. The resulting light-adaptive texture features are fused into the main branch of the model, enriching its representation capability and contributing to performance gains. Importantly, this mechanism requires adjusting the pupil dilation factor $α$ suitable for well-lit scenes, similar to how human vision experiences light adaptation after exiting a dark tunnel. Tuning the pupil dilation factor α allows the algorithm to adapt rapidly to changing illumination conditions—just as the biological pupil would. Ultimately, this bio-inspired mechanism provides a cost-efficient and adaptive solution to low-light compensation, leveraging the inherent environmental adaptability of biological vision systems, which is superior to conventional image-centric enhancement approaches in perception task.

### 3.3 Biologically inspired separable representation learning strategy
#### 3.3.1 Overall framework of the SLVM

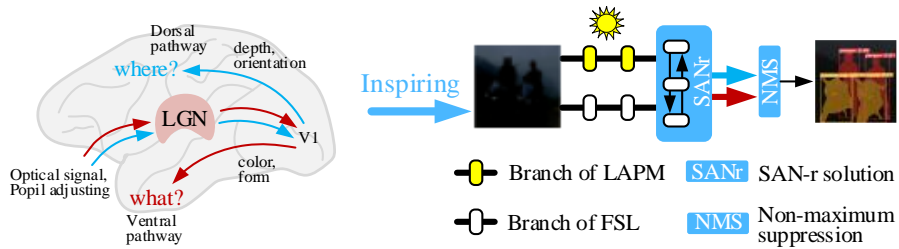

**Figure 6.** The SLVM framework inspired by primate visual systems. The proposed architecture consists of three biologically inspired components: the Branch of LAPM, responsible for luminance-aware feature modulation through the Light Adaptive Pupillary Mechanism; the Branch of FSL, which implements feature-level separable learning in the backbone for efficient and expressive representation learning; and the SANr, a secondary feature alignment solution based on the SNI-r approach, refining spatial features via pixel-wise adaptive modulation.

Building upon three key inspirations from biological vision systems, we propose a biologically inspired visual perception framework - the SLVM - tailored for object understanding in low-light transportation scenes. As illustrated in Figure. 6, the model architecture is designed to be modular, scalable, end-to-end trainable, and real-time deployable. The framework integrates the following biologically motivated mechanisms: (1) Pupil-inspired adaptive illumination modulation, which enhances the perception of object textures under poor lighting by simulating the pupil's contraction and dilation behavior. The photosensitive mask of the LAPM module enables robust feature extraction across diverse illumination conditions after tuning the pupil scaling factor based on scene brightness; (2) Feature-level separable learning, implemented via the proposed FSLConv module, not only reduces computational complexity but also improves the expressiveness of learned representations through biologically plausible branching; (3) Task-specific decoupled learning, which mirrors the biological strategy of assigning different perceptual objectives (e.g., object classification, spatial localization, and luminance adaptation) to distinct neural circuits. By dedicating specialized branches (e.g., LAPM for luminance-sensitive perception, classification and regression heads for recognition and localization), the SLVM avoids mutual interference and facilitates more efficient learning on each sub-task.

Together, these components form a cohesive and elegant perception pipeline. To accommodate varying computational constraints, we design two model variants of SLVM by adjusting network width and depth: SLVM-S (231 layers, 3.33M parameters, 11.8 GFLOPs @ 640×640) and SLVM-L (331 layers, 80.4M parameters, 244.1 GFLOPs @ 640×640). Detailed architectural specifications and implementation code are available at https://github.com/alanli1997/slvm.

### 3.3.2 Feature-level separability and vision-task separability in the SLVM

Modern deep learning-based perception models aim to learn compact and effective representations from high-dimensional raw image data. This transformation—from raw sensory input to abstract representations—is conceptually analogous to the process through which intelligent biological organisms convert external instructions into actions. For example, in the training process of service dogs, the transition from a regular dog to a fully functional service dog involves multiple incremental learning steps, rather than a single end-to-end training phase [8], as illustrated in Figure. 7(a). Similarly, in visual perception models, an initially untrained network gradually learns to map images to meaningful representations and outputs—mimicking how biological entities are incrementally conditioned. This leads to the second core inspiration from biological learning systems: stepwise representation learning, illustrated in Figure. 7(b).

Rather than learning complex tasks holistically, biological systems often decompose them into a sequence of sub-tasks. Such decomposition not only reduces cognitive load but also improves learning efficiency. Likewise, in vision models, multi-stage feature extraction (e.g., via feature pyramids) has become a standard approach, replacing monolithic convolutional or transformer blocks with hierarchically structured stages (e.g., Path 2 in Figure. 7(b)).

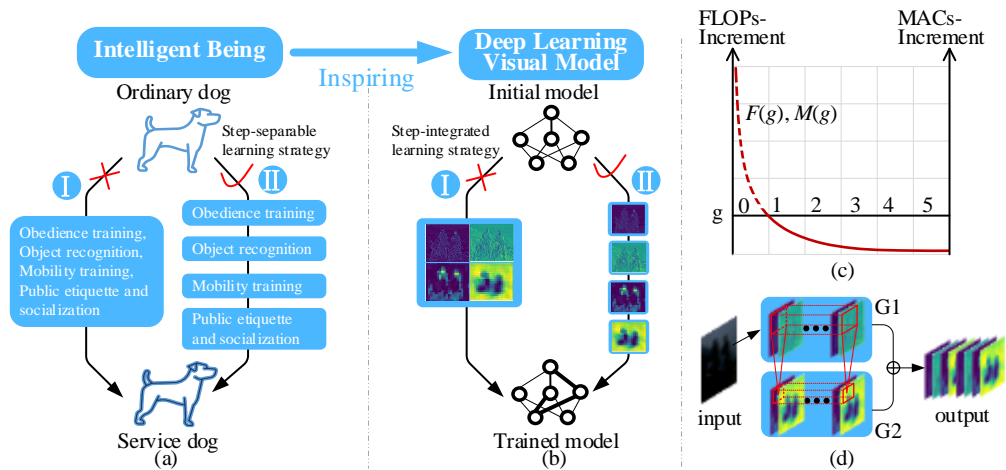

**Figure 7.** Biologically inspired separable representation learning mechanism derived from the training behavior of service dogs. (a) Decomposition of the service dog learning process. (b) Biologically inspired separable learning process for visual models. (c) Curves of incremental functions F(g) and M(g), used to analyze and determine the optimal number of feature splits. (d) Schematic illustration of the proposed FSLConv structure.

However, we argue that coarse stage-level decomposition can be further refined into feature-level decomposition—which we define as the first requirement of separable learning. This fine-grained decomposition logic is further supported by insights from fractal theory [26], a field of natural science that explains how complexity can emerge from the recursive application of simple rules. Although modern architectures such as ResNet [27] and ViT [28] do not explicitly reference fractals, they implicitly reflect fractal characteristics, such as recursive structure reuse and iterative processing. Motivated by this, we propose to leverage fractal-inspired convolutional operations, such as repeated convolutions with consistent kernel sizes, to build a stepwise feature refinement module. However, such structural decomposition inevitably introduces a trade-off between representational richness and computational complexity.

Deeper or more branched networks often exhibit better representational capacity, yet they may suffer from higher computational and memory overhead. Conversely, simpler networks facilitate smoother information flow and lower latency but often exhibit poor parameter utilization—a key limitation in legacy models like VGG [29] or even ResNet when applied to fine-grained recognition tasks [30]. At the other extreme, architectures like ViT rely on dense self-attention, where every token attends to every other, yielding exceptional modeling capacity but abandoning inductive biases such as locality and translation invariance. This results in significantly higher training and inference costs, making such models impractical for real-time tasks, especially on resource-constrained edge devices.

Hence, a central question emerges: What is the optimal level of branching? To answer this, we turn to the two most fundamental model performance metrics: floating-point operations (FLOPs) and memory access cost (MACs). In the following, we derive analytical formulas for FLOPs and MACs of standard and grouped convolutional layers, and introduce their incremental functions with respect to the number of groups $g$, offering a quantitative basis for evaluating and selecting appropriate feature partitioning strategies. Let the input and output feature maps be of dimensions $C_1 \times H \times W$ and $C_2 \times H \times W$, respectively. For a convolution kernel of size $K_h \times K_w$, the FLOPs of a standard convolution operation can be expressed as Equation (4):

$$FLOPs = [C_1 K_h K_w + (C_1 K_h K_w - 1) + 1] C_2 HW = 2 C_1 C_2 K_h K_w HW \quad (4)$$

For a convolution operation with split factor $g$, each group handles $C_1/g$ input channels and $C_2/g$ output channels. The $FLOPs_g$ is given by Equation (5):

$$FLOPs_g = \left[\frac{C_1 K_h K_w}{g} + \left(\frac{C_1 K_h K_w}{g} - 1\right) + 1\right] \frac{C_2}{g} HW = 2 \frac{C_2}{g} C_1 K_h K_w HW \quad (5)$$

The FLOPs increment, F(g), with respect to split factor $g$ can be formulated as Equation (6):

$$F(g) = FLOPs_g - FLOPs = 2\left(\frac{1-g}{g}\right) C_1 C_2 K_h K_w HW \quad (6)$$

Similarly, the MACs increment, M(g), is derived in Equation (7):

$$M(g) = MACs_g - MACs = C_1 C_2 K_h K_w \left(\frac{1}{g} - 1\right) \quad (7)$$

Here, the MACs of the standard and the split are defined in Equation (8) and (9), respectively:

$$MACs = HW(C_1 + C_2) + C_1 C_2 K_h K_w \quad (8)$$

$$MACs_g = \left[HW\left(\frac{C_1}{g} + \frac{C_2}{g}\right) + \frac{C_1}{g} \frac{C_2}{g} K_h K_w\right] g \quad (9)$$

We observe that both F(g) and M(g) form segments of hyperbolic functions, as illustrated in Figure 6(c). As the split number $g$ increases, the rate of improvement in FLOPs and MACs gradually diminishes until reaching saturation. Within the defined domain, the point of maximum return occurs at $g=1$, followed by $g=2$. Therefore, when considering hardware factors such as data synchronization, $g=2$ represents the optimal cost-performance ratio, where both F(g) and M(g) achieve a reduction rate of 1/4. This configuration ($g=2$) aligns with biological branching patterns and constitutes the simplest bifurcation scheme—we designate this as Requirement 2: limiting the feature splitting degree to 2. Furthermore, to maximize utilization of GPU parallel computing capabilities while maintaining model capacity through parameter efficiency, we establish Requirement 3: maintaining symmetric input-output channel dimensions. By simultaneously satisfying Requirements 1-3, we propose the FSLConv, whose architecture is depicted in Figure 7(d). The computational procedure is detailed as follows:

Given an input feature tensor $X \in \mathbb{R}^{C_1 \times H \times W}$, the FSLConv produces an output $Y \in \mathbb{R}^{C_2 \times H \times W}$ through a two-stage separable convolutional process. The first-stage feature map $G_1 \in \mathbb{R}^{C_o \times H \times W}$ is computed using convolution weights $W_1 \in \mathbb{R}^{C_o \times C_1 \times 3 \times 3}$, where $C_o = C_2/2$. The feature computation includes normalization and gated activation, as Equation (10):

$$G_1 = Act\left(\frac{\frac{W_1 \cdot X - E(W_1 \cdot X)}{\sqrt{Var(W_1 \cdot X) + \epsilon}} \cdot \gamma + \beta}{1 + e^{-\left[\frac{W_1 \cdot X - E(W_1 \cdot X)}{\sqrt{Var(W_1 \cdot X) + \epsilon}} \cdot \gamma + \beta\right]}}\right) \quad (10)$$

where Act(·) denotes a nonlinear activation function (e.g., ReLU or SiLU), E(·) denotes the mean value and Var(·) denotes the variance, both computed across the batch and spatial dimensions for batch normalization, $\epsilon = 10^{-8}$ is a small constant to ensure numerical stability, and $\gamma$ and $\beta$ are learnable batch normalization scaling and shifting parameters.

The second-stage feature map $G_2 \in \mathbb{R}^{C_o \times H \times W}$ is produced using a second convolution layer $W_2 \in \mathbb{R}^{C_o \times C_o \times 3 \times 3}$ also followed by normalization and gated activation, as Equation (11):

$$G_2 = Act\left(\frac{\frac{W_2 \cdot G_1 - E(W_2 \cdot G_1)}{\sqrt{Var(W_2 \cdot G_1) + \epsilon}} \cdot \gamma + \beta}{1 + e^{-\left[\frac{W_2 \cdot G_1 - E(W_2 \cdot G_1)}{\sqrt{Var(W_2 \cdot G_1) + \epsilon}} \cdot \gamma + \beta\right]}}\right) \quad (11)$$

Finally, the two decomposed features are concatenated along the channel dimension to form the output:
$$Y = G_1 \oplus G_2$$
where $\oplus$ denotes concatenation along the channel axis.

The third inspiration from natural intelligent biological visual systems—perceptual task decomposition learning—is drawn from the visual processing mechanisms of primates [31, 32]. Through long-term evolution and competitive selection, biological vision has developed optimized pathways for data transmission by specializing different visual tasks and delegating them to dedicated neural branches. These include spatial localization and motion estimation, object shape and category recognition, and adaptation to varying lighting conditions. Interestingly, even biologically agnostic designs in machine vision have empirically demonstrated that differentiated processing of visual cues via structural separation often yields superior performance [33, 34], aligning closely with principles observed in biological vision systems.

In experiments, we attempted to inject location cues directly into the regression branch and apply spatial feature enhancement on the classification branch. However, we observed a slight performance drop under such settings. In contrast, when we introduced the LAPM module as an independent branch dedicated to perceiving illumination-sensitive cues in low-light scenarios, we observed a consistent improvement in performance. We hypothesize that this phenomenon stems from the fact that high-level abstract features are vulnerable to contamination when strong cues are forcibly injected. Such aggressive injection may distort finely learned representations, akin to feature misalignment issue caused by coarse fusion in FPN-like architectures [21]. In contrast, the independent branch architecture adopted by LAPM allows for gradual, gentle learning of illumination-aware features, resulting in more stable adaptation.

### 3.3.3 The SNI-r for multi-scale feature spatial alignment fusion in the SLVM

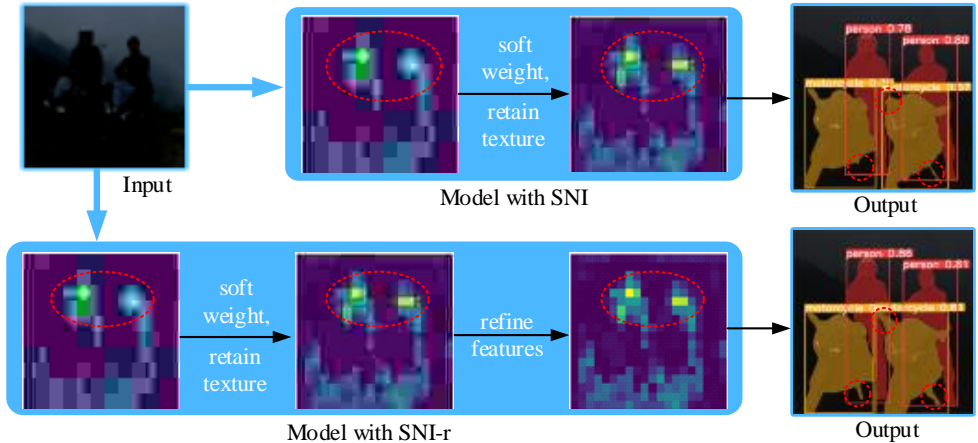

**Figure 8.** Qualitative comparison of texture features under SNI and SNI-r. Visualization of texture feature representations produced by the SNI module and the proposed SNI-r module. The SNI-r demonstrates enhanced spatial texture alignment and improved feature consistency in challenging low-light scenes. And the model with SNI-r achieved higher instance segmentation accuracy, as indicated by the red dashed circle in the output.

To address misalignment issue in FPN-like architectures, Li et al. [21] proposed the Secondary Feature Alignment Fusion Neck (SAN) solution, a cost-free method that alleviates the abruptness of high-to-low feature fusion by softening the influence of high-level features. However, we observe that the SAN operates at the patch-level, approximating a global optimization. This global bias may suppress the expressiveness of inherently weighted high-level features. To overcome this limitation, we introduce SNI-r, a spatially-aware refinement of the SAN strategy that performs pixel-scale alignment across feature maps. Specifically, we retain the soft-weights mechanism of SNI, while applying adaptive linear mapping to each spatial feature point, which is then normalized via a sigmoid function to represent its relative importance as a probabilistic spatial density. These weights are used to modulate the original features in a locally adaptive yet globally non-competitive manner. Let $X \in \mathbb{R}^{C_1 \times H \times W}$ denote the input and $Y \in \mathbb{R}^{C_2 \times 2H \times 2W}$ denote the corresponding output. The refined feature adjustment via SNI-r is formulated as Equation (12):
$$Y = (\alpha \cdot \mu(X)) \odot \sigma(W * (\alpha \cdot \mu(X)) + b) \quad (12)$$
where α is a resolution compensation coefficient (equal to the square of the inverse upsampling scale); $\mu(\cdot)$

denotes nearest-neighbor upsampling; * represents the convolution operation; $W \in \mathbb{R}^{C_2 \times C_1 \times 1 \times 1}$ and $b \in \mathbb{R}^{C_2}$ are the trainable weights and bias of the 1×1 convolution; $\sigma(\cdot)$ denotes the sigmoid function; and $\odot$ indicates element-wise (Hadamard) product.

This localized probabilistic gating enhances spatial adaptability while avoiding dominance conflicts across the global feature space. A qualitative visualization of feature refinement is shown in Figure. 8, where the spatial density adjustment facilitates improved contextual understanding under challenging conditions.

## 4. Experiments

In this section, we present detailed ablation studies and comparisons with state-of-the-art real-time object perception models. All experiments were conducted on a workstation equipped with an NVIDIA Tesla A40 GPU and an Intel® Xeon® Gold 6338 CPU @ 2.00GHz × 24, running Ubuntu 20.04. For static perception tasks (object detection and instance segmentation), the models were trained for 200 epochs with a batch size of 16 and an input resolution of 640. We used the SGD optimizer with a learning rate of 0.0001, momentum of 0.937, and a weight decay of 0.0005. For dynamic perception tasks (optical flow), models were trained for 10k steps with the AdamW optimizer. The input resolution was 320×1152, and we set the learning rate to 0.0002 with a weight decay of 0.0001. More implementation details and training configurations are reported in supplementary materials.

### 4.1 Dataset

Object perception in dark transportation scenes is a critical challenge for enabling high-level autonomous driving and navigation. Most existing perception models rely on static sensing paradigms, focusing only on spatial localization and object classification, while ignoring motion cues. In vision-based frameworks, optical flow estimation serves as a well-established approach for motion perception. However, estimating motion under low-light conditions remains highly challenging due to the significant reduction of photon counts, which severely impairs the ability to capture motion dynamics and leads to pronounced motion blur. To address this issue, we extend the conventional static perception framework to include dynamic perception in low-light scenarios, with a focus on optical flow prediction under illumination degradation. We evaluate our model on three benchmark datasets tailored to dark transportation environments: LIS-traffic, Dark-traffic, and the original LIS dataset. The tasks include object detection, instance segmentation, and optical flow estimation. For rapid ablation studies, we use LIS-traffic, a real-world low-light dataset containing 1,547 low-light images and 6,359 pixel-level instance annotations. Captured under short exposure settings, this dataset reflects authentic dark conditions and allows fast, cost-efficient experimentation, contributing to lower carbon emissions and environmental sustainability. However, its limited size may lead to performance bias and reduce the reliability of model comparisons. To overcome this limitation, we conduct competitive evaluations on the large-scale Dark-traffic benchmark—the most extensive instance segmentation dataset for dark scenes to date—with 10,425 low-light images and 99,014 pixel-level instance masks. For optical flow estimation under degraded illumination, we adopt the degraded version of KITTI [35], consisting of 200 image pairs with ground-truth flow annotations. Finally, to ensure fair comparisons with illumination enhancement methods, we also include evaluations on the original LIS benchmark, where our model competes against state-of-the-art low-light perception frameworks. The Dark-traffic dataset is available at:
https://drive.google.com/drive/folders/1B8EzDn64bGBgyRCfppL_jhcOA3hIwnzi.

### 4.2 Ablation study

#### 4.2.1 Effect of illumination degradation on general object perception models

We first evaluate the impact of lighting conditions on the performance of state-of-the-art real-time object perception models. Results are summarized in Table 2. A consistent and significant performance drop is observed when transitioning from well-lit to low-light scenarios across both the LIS-traffic dataset, which contains real-world low-light samples, and the COCO-traffic dataset, which is generated through illumination transfer. All tested models exhibit a substantial decline in accuracy, with metrics decreasing by nearly 10 percentage points on average, and up to 13 percentage points in the worst case. These findings indirectly validate the physical interpretability of real-data-based illumination transfer algorithms, as well as the inherent limitation of general-purpose perception models under low-light conditions. Specifically, these models fail to reliably extract foreground features from degraded and chaotic visual input in dark environments. Comparing with the biological visual system, these existing models show significant limitations: lacking adaptive pupillary mechanisms, their visual intake remains constant regardless of ambient light changes, severely constraining their scene-level robustness in dim environments.

**Table 2**. Ablation study on baseline (YOLOv8-Nano: 3.24 million parameters, 11.1 GFLOPs@640×640) performance under well-lit and low-light conditions on the LIS-traffic and COCO-traffic datasets. All metrics show significant degradation under low-light settings.

| Dataset | Light condition | Latency$^{b=16}_{(ms)}$ | $AP^{box}_{50}$(%) | $AP^{box}$(%) | $AP^{mask}_{50}$(%) | $AP^{mask}$(%) |
|---|---|---|---|---|---|---|
| LIS-traffic | well-lit | 2.4 | 65.7 | 49.1 | 62.4 | 42.9 |
|  | low-lit | 2.5 | 57.0(↓8.7) | 39.5(↓9.6) | 51.6(↓10.8) | 32.7(↓10.2) |
| COCO-traffic | well-lit | 2.6 | 55.2 | 34.7 | 51.6 | 28.1 |
|  | low-lit | 2.6 | 42.2(↓13.0) | 26.8(↓7.9) | 40.1(↓11.5) | 20.7(↓7.4) |

**4.2.2 Impact of serial feature splitting in the FSLConv**

Next, we perform an ablation study on Requirement 2 of the proposed FSLConv module—the number of serial feature splits—as shown in Table 3. To better leverage GPU parallelism, we adopt group sizes that are powers of two, evaluating representative values of $g \in \{1, 2, 4, c\}$, where g denotes the number of serial splits: g = 1 corresponds to conventional holistic learning, without any feature-level splitting; g = 2 implements a two-way serial separable learning scheme—essentially a single recursive decomposition step (analogous to two total iterations in fractal theory, including the original); g = 4 applies a four-way serial splitting strategy (corresponding to four recursive iterations in fractal terms); g = c represents a channel-wise full decomposition (i.e., one feature per group), achieving the maximum number of recursive iterations possible.

**Table 3**. Ablation study of different feature split settings for the FSLConv module on the LIS-traffic. The setting g = 2 achieves the best trade-off between accuracy and efficiency, while g = c results in the poorest performance.

| Groups | Param(M) | FLOPs(G) | Latency$^{b=16}_{(ms)}$ | $AP^{box}_{50}$(%) | $AP^{box}$(%) | $AP^{mask}_{50}$(%) | $AP^{mask}$(%) |
|---|---|---|---|---|---|---|---|
| g=1 (standard) | 3.24 | 11.1 | 2.5 | 57.0 | 39.5 | 51.6 | 32.7 |
| g=2 | 3.19 | 11.1 | 2.5 | **57.2(↑0.2)** | **40.3(↑0.8)** | **52.5(↑0.9)** | **34.1(↑1.4)** |
| g=4 | 3.10 | 10.9 | 2.6 | 56.8(↓0.2) | 39.7(↑0.2) | 51.4(↓0.2) | 32.7(-) |
| g=c | 2.86 | 10.4 | 2.9 | 56.4(↓0.6) | 38.8(↓0.7) | 51.4(↓8.7) | 32.4(↓0.3) |

The results reveal that *g* = 2 yields the best overall trade-off, achieving consistent accuracy improvements while preserving computational efficiency. In contrast, excessive decomposition (*g* = c) leads to degraded performance due to feature fragmentation and loss of global context. As discussed in Section 3.3, this outcome is aligned with the unfavorable cost-benefit tradeoff described by the capacity-overhead functions F(g) and M(g). Moreover, extreme splitting introduces gradient instability and information loss [37], making *g* = c an undesirable setting despite its theoretical maximal decomposition.

**4.2.3 Effectiveness of the SNI-r vs. the SNI**

**Table 4.** Comparison between SNI and SNI-r on LIS-traffic and Dark-traffic.

| Model | Param(M) | FLOPs(G) | Latency$^{b=16}_{(ms)}$ | $AP^{box}$(%) | $AP^{mask}$(%) | $AP^{mask}_{50}$(%) | $AP^{mask}_{75}$(%) |
|---|---|---|---|---|---|---|---|
| \multicolumn{8}{c}{LIS-traffic} ||||||||
| baseline | 3.24 | 11.1 | 2.5 | 39.5 | 32.7 | 51.6 | 32.7 |
| +SNI | 3.24 | 11.1 | 2.4 | 40.2(↑0.7) | 33.0(↑0.3) | 52.6(↑1.0) | **34.7(↑2.0)** |
| +SNI-r | 3.33 | 11.6 | 2.8 | 40.2(↑0.7) | **33.2(↑0.5)** | **52.7(↑1.1)** | 33.6(↑0.9) |
| \multicolumn{8}{c}{Dark-traffic} ||||||||
| baseline | 3.24 | 11.1 | 3.1 | 28.8 | 23.1 | 41.4 | 18.6 |
| +SNI | 3.24 | 11.1 | 3.1 | 36.3(↑7.5) | 24.8(↑1.7) | 45.5(↑4.1) | 22.9(↑4.3) |
| +SNI-r | 3.33 | 11.6 | 3.2 | **36.5(↑7.7)** | **24.9(↑1.8)** | **46.4(↑5.0)** | **23.1(↑4.5)** |

We further evaluate the proposed SNI-r module against the original SNI (Table 4). The results show only marginal improvements for SNI-r over SNI in this experiment, which is expected under mild feature misalignment. Original SNI operates directly without additional learning, making it lightweight and effective when the spatial receptive field mismatch across fused features remains small. However, when integrating more heterogeneous features (e.g., light-adaptive textures or binary priors), the enhanced alignment capability of SNI-r becomes crucial. By introducing a learnable linear mapping, SNI-r enables pixel-wise modulation of spatial attention, better compensating for alignment discrepancies that become more prominent under extreme low-light conditions.

**4.2.4 Comprehensive analysis of module contributions**

We conduct a detailed ablation study of the proposed LAPM, FSLConv, and SNI-r modules on the LIS-traffic dataset under real-world low-light conditions. As reported in Table 5, all three components demonstrate consistent effectiveness, confirming their individual and combined contributions to model performance. Among the three, FSLConv—inspired by the core principle of biologically grounded separable learning—emerges as the most generally effective standalone module. It improves all prediction metrics by an average of 1.225 percentage points, showcasing its adaptability across varied low-light scenarios. The least individually impressive improvement is observed from LAPM, which yields only 0.1–0.2 percentage points of gain in isolation. However, this does not imply inefficacy. LAPM focuses on light-sensitive texture modulation, which tends to produce sharper, high-frequency feature boundaries. These are inherently more difficult for the model to fuse with conventional textures, leading to competitive gradient flow and training instability unless guided by well-aligned feature fusion mechanisms. Crucially, the best-performing configurations are: FSLConv + SNI-r, and FSLConv + SNI-r + LAPM. The latter outperforms the former by +1.0 $AP^{box}$, +1.6 $AP^{mask}$, and +1.4 $AP^{mask}_{50}$, clearly demonstrating that LAPM can be effectively harnessed when paired with robust learning and fusion strategies.

**Table 5.** Combined ablation results for LAPM, FSLConv, and SNI-r on LIS-traffic. FSLConv provides the highest standalone improvement, while LAPM shows its full potential when integrated with proper fusion mechanisms.

| Module | Param(M) | FLOPs(G) | Latency$^{b=16}_{(ms)}$ | $AP^{box}$(%) | $AP^{mask}$(%) | $AP^{mask}_{50}$(%) | $AP^{mask}_{75}$(%) |
|---|---|---|---|---|---|---|---|
| baseline | 3.24 | 11.1 | 2.5 | 39.5 | 32.7 | 51.6 | 32.7 |
| LAPM | 3.24 | 11.1 | 2.5 | 39.6(↑0.1) | 32.9(↑0.2) | 51.8(↑0.2) | 32.6(↓0.1) |
| FSLConv | 3.19 | 11.1 | 2.5 | 40.3(↑0.8) | 34.1(↑1.4) | 52.5(↑0.9) | 34.5(↑1.8) |
| SNI-r | 3.33 | 11.6 | 2.8 | 40.2(↑0.7) | 33.2(↑0.5) | 52.7(↑1.1) | 33.6(↑0.9) |
| LAPM, SNI-r | 3.32 | 11.6 | 2.8 | 40.6(↑1.1) | 33.0(↑0.3) | 52.2(↑0.6) | 34.4(↑1.7) |
| LAPM, FSLConv | 3.24 | 11.1 | 2.6 | 41.2(↑1.7) | 33.6(↑0.9) | 52.7(↑1.1) | 33.2(↑0.5) |
| FSLConv, SNI-r | 3.32 | 11.5 | 2.8 | 40.6(↑1.1) | 33.5(↑0.8) | 53.0(↑1.4) | 34.9(↑2.2) |
| FSLConv, SNI-r, LAPM | 3.32 | 11.8 | 2.8 | **41.6(↑2.1)** | **35.1(↑2.4)** | **54.4(↑2.8)** | **35.6(↑2.9)** |

**4.2.5 Generalization to low-light optical flow estimation**

To evaluate the potential of our biologically inspired SLVM framework in dynamic perception tasks, we integrate it into two representative optical flow estimation models: the GMFlow [51] and the real-time NeuFlow2 [53]. We assess its performance under low-light conditions on the Dark-traffic benchmark. The quantitative and qualitative results are presented in Table 6 and Figure. 10, respectively. Notably, the fast-adaptive LAPM module reduces the EPE of GMFlow and NeuFlow2 by 3.73% and 12.37%, respectively, while consistently maintaining lower inference latency. These results demonstrate that our illumination-adaptive mechanism generalizes beyond recognition tasks and is also effective for dense prediction under degraded lighting conditions. Specifically, in low-light scenes where motion blur and texture suppression degrade standard optical flow performance, our approach provides enhanced robustness.

This experiment motivates our next research direction: extending object perception in low-light traffic scenarios from static recognition to dynamic modeling by incorporating spatial motion understanding. It is important to emphasize that most current real-time perception models offer only pseudo-dynamic understanding—they rely on frame-by-frame detection and post hoc association, rather than explicitly modeling object motion. During inference, such models do not perceive actual motion trajectories. In future work, we aim to develop a real-time perception framework capable of not only detecting and segmenting traffic objects under low-light conditions, but also estimating their true motion in space. This will enable physically grounded understanding of dynamic scenes. Achieving this goal requires tightly coupling real-time object perception with real-time optical flow or scene flow estimation, opening up new possibilities for physically consistent modeling in complex, low-light traffic environments.

Table 6. Ablation results of LAPM on low-light optical flow estimation (input size is 320×1152).

| Model | Param(M) | FLOPs(G) | Latency$_{(ms)}^{b=1}$ | EPE |
|---|---|---|---|---|
| KITTI (well-lit) | | | | |
| GMFlow [51] | 2.07 | 225.0 | 238.5 | 3.519 |
| NeuFlow2 [53] | 9.03 | 128.3 | 85.0 | 1.511 |
| Dark-traffic (low-lit) | | | | |
| GMFlow [51] | 2.07 | 225.0 | 239.9 | 4.748 |
| NeuFlow2[53] | 9.03 | 128.3 | 85.9 | 2.417 |
| **GMFlow+LAPM(ours)** | 2.08 | 225.1 | 214.7 | **4.571(↓3.73%)** |
| **NeuFlow2+LAPM(ours)** | 9.03 | 128.3 | 85.2 | **2.118(↓12.37%)** |

### 4.3 Comparative experiment

To evaluate the effectiveness of our proposed framework, we compare it against several state-of-the-art real-time perception models, including YOLACT [7], YOLOv8 [38], YOLOv9 [39], YOLOv10 [40], YOLOv11 [41], YOLOv12 [42], and RT-DETR [43]. The experiments focus on two core visual tasks: object detection and instance segmentation under low-light conditions.

Table 7. Comparative instance segmentation results on the Dark-traffic with state-of-the-art real-time models. (* denotes ResNet-18 backbone with PANet.)

| Model | Param(M) | FLOPs(G) | Latency$_{(ms)}^{b=16}$ | $AP^{mask}$(%) | $AP_{50}^{mask}$(%) | $AP_{75}^{mask}$(%) |
|---|---|---|---|---|---|---|
| YOLACT*[7] | 10.76 | 59.6 | 16.7 | 20.6 | 39.4 | 18.4 |
| YOLOv8n[38] | 3.24 | 11.1 | 3.1 | 23.1 | 41.4 | 18.6 |
| YOLOv9t[39] | 2.18 | 11.4 | 4.5 | 23.4 | 41.4 | 19.3 |
| YOLO11n[41] | 2.84 | 10.2 | 2.9 | 23.1 | 41.0 | 18.8 |
| Yolo12n[42] | 2.81 | 10.2 | 3.7 | 22.6 | 40.5 | 18.4 |
| **SLVM-S(ours)** | 3.32 | 11.8 | 2.4 | **28.7** | **50.7** | **23.2** |

Table 8. Comparative object detection results on the Dark-traffic with state-of-the-art real-time models.

| Model | Param(M) | FLOPs(G) | Latency$_{(ms)}^{b=16}$ | $AP^{box}$(%) | $AP_{50}^{box}$(%) | $AP_{75}^{box}$(%) |
|---|---|---|---|---|---|---|
| YOLOv8n[38] | 3.01 | 8.1 | 10.0 | 28.8 | 46.3 | 37.1 |
| YOLOv9t[39] | 1.97 | 7.6 | 9.5 | 28.0 | 45.0 | 33.8 |
| YOLOv10n[40] | 2.70 | 8.2 | 10.2 | 27.7 | 44.1 | 34.0 |
| YOLO11n[41] | 2.58 | 6.3 | 9.2 | 27.9 | 45.0 | 34.2 |
| YOLOv12n[42] | 2.56 | 6.3 | 9.6 | 28.7 | 45.6 | 35.0 |
| RT-DETR[43] | 10.43 | 24.8 | 15.5 | 24.7 | 40.5 | 29.1 |
| **SLVM-S(ours)** | 3.09 | 8.5 | 6.5 | **35.9** | **55.4** | **37.4** |

Table 7 presents the instance segmentation performance of our proposed SLVM-S model on the Dark-traffic benchmark, alongside competitive baselines. As shown, SLVM-S significantly outperforms all comparison models in segmentation accuracy. It achieves an average improvement of over 5 percentage points in $AP^{mask}$, and nearly 10 percentage points in $AP_{50}^{mask}$ over the best-performing baselines. Table 8 reports the object detection performance of SLVM-S compared to the most advanced real-time detection models. SLVM-S again demonstrates superiority, surpassing RT-DETR—which leverages a Transformer-based architecture—by 11.2 percentage points in AP, while consuming only 34.27% of its computational cost.

To enable a more equitable comparison with image enhancement-based approaches, we conducted competitive evaluations against recent state-of-the-art methods and large-scale visual models on the publicly available LIS dataset; results are presented in Table 9. Notably, we report both FLOPs and inference latency (with batch size = 1) for all models to ensure a fair and comprehensive performance assessment, as computational efficiency is critical in real-time perception systems. SLVM-L consistently outperformed all competing models across all metrics while maintaining lower computational cost and faster inference time. Without relying on any illumination enhancement or denoising (E & D) techniques, SLVM-L surpassed the large CNN-based model ConvNeXt-T [45] by 21.9 percentage points in $AP^{box}$ and outperformed the Transformer-based Swin-T by 25 percentage points. Even compared to fine-tuned models like Mask-RCNN

[44] on the LIS benchmark, SLVM-L still demonstrated superior performance with a lower FLOPs count (244.1G vs. 280.0G) and reduced latency (17.6 ms vs. 17.8 ms). In cases where competing models adopted advanced pre-processing techniques such as illumination enhancement or denoising, SLVM-L—trained end-to-end without such auxiliary strategies—still achieved the best performance. For instance, SLVM-L exceeded Swin-T [46] + EnlightenGAN [22] and ConvNeXt-T + EnlightenGAN by an average margin of 11.125 percentage points across all precision metrics, while maintaining lower computational cost. Moreover, it outperformed REDI [49] + Mask-RCNN by an average of 3.1 percentage points, despite the latter's use of both enhancement and fine-tuning. These results highlight SLVM-L's balance between task performance and computational efficiency, demonstrating its practicality for real-time deployment in resource-constrained environments.

**Table 9.** Comparison with more SOTA vision models (including those utilizing illumination enhancement and denoising techniques) on the original LIS benchmark. † indicates the model has been fine-tuned on the LIS dataset; see [1] for fine-tuning details.

| Model | E & D | FLOPs(G) | Latency$_{(ms)}^{b=1}$ | AP$^{box}$(%) | AP$^{mask}$(%) | AP$_{50}^{mask}$(%) | AP$_{75}^{mask}$(%) |
|---|---|---|---|---|---|---|---|
| MaskRCNN†[1,44] | ✗ | 280.0 | 17.8 | 42.9 | 35.5 | 57.5 | 36.1 |
| ConvNeXt-T [45] | ✗ | 266.7 | 33.3 | 27.9 | 23.7 | 41.4 | 23.7 |
| Swin-T [46] | ✗ | 267.1 | 39.2 | 24.8 | 20.7 | 38.2 | 19.5 |
| Mask2Former[47] | ✗ | 293.0 | 43.1 | 22.9 | 21.4 | 37.9 | 20.9 |
| PointRend [48] | ✗ | 260.0 | 39.8 | 23.5 | 20.6 | 37.2 | 19.2 |
| MaskRCNN†[1,44] | REDI [49] | - | 22.7 | 42.8 | 36.0 | **59.0** | 35.8 |
| ConvNeXt-T [45] | REDI [49] | - | 28.5 | 32.2 | 27.6 | 46.6 | 27.5 |
| Swin-T [46] | REDI [49] | - | 32.1 | 29.8 | 25.7 | 43.7 | 25.3 |
| Mask2Former [47] | REDI [49] | - | 34.4 | 27.7 | 24.0 | 42.2 | 23.2 |
| PointRend [48] | REDI [49] | - | 35.3 | 28.1 | 26.7 | 44.1 | 26.0 |
| Swin-T [46] | EnlightenGAN[22] | - | 27.5 | 33.2 | 27.8 | 48.4 | 28.3 |
| ConvNeXt-T[45] | EnlightenGAN[22] | - | 25.3 | 35.9 | 29.5 | 50.8 | 29.1 |
| **SLVM-L(ours)** | ✗ | 244.1 | 17.6 | **49.8** | **38.3** | 58.4 | **39.5** |

### 4.4 Visualization

To further demonstrate the effectiveness of the proposed SLVM framework, we present qualitative comparisons on both static and dynamic perception tasks under low-light conditions. For static perception tasks such as object detection and instance segmentation, we compare SLVM against several real-time baselines, including the YOLO series (YOLOv8-N, YOLOv10-N, YOLOv12-N) and RT-DETR. As shown in Figure 9, RT-DETR exhibits the weakest performance, with a considerable number of false positive detections. In contrast, the YOLO series and SLVM produce predictions that are much closer to the ground truth. Notably, SLVM consistently yields the most precise segmentation masks. For instance, in the area marked with a red dashed circle in Figure. 9, only SLVM successfully identifies the subtle structure of the car's side mirror, highlighting its superior fine-grained perception capability.

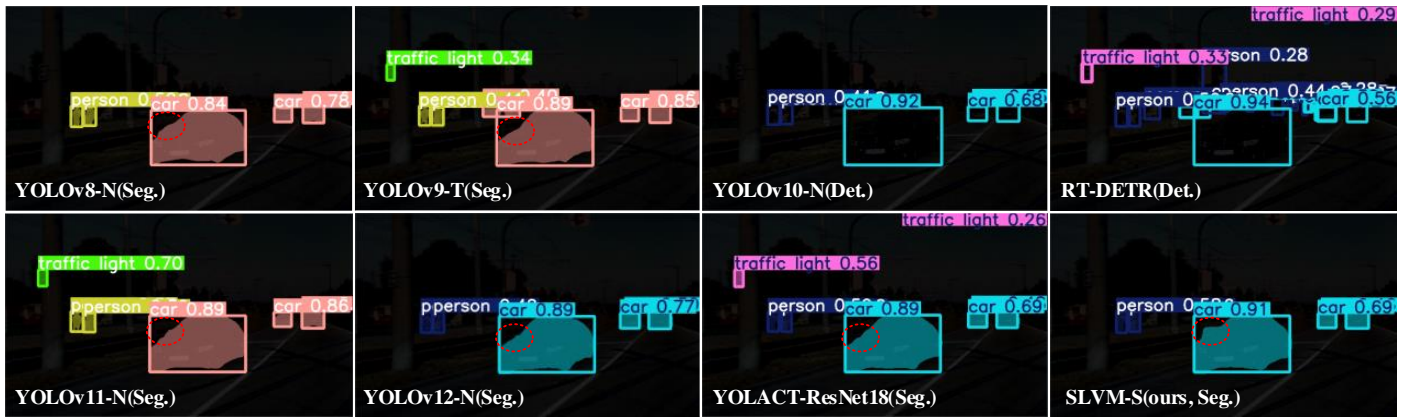

**Figure 9.** Qualitative comparisons on static vision tasks (object detection and instance segmentation). In low-light traffic scenes, SLVM produces the fewest false predictions and the most accurate mask boundaries.

For dynamic perception tasks, we evaluate optical flow estimation in low-light traffic scenarios, comparing SLVM-integrated versions of GMFlow and NeuFlow2 against their original counterparts. The inclusion of LAPM leads to significant improvements: the EPE is reduced by 3.73% for GMFlow and by 12.37% for NeuFlow2. These quantitative gains are clearly reflected in the visual results as illustrated in Figure 10. The SLVM-enhanced models exhibit sharper and more coherent motion boundaries compared to their baselines. Among them, NeuFlow2+LAPM achieves the lowest EPE (2.118) and delivers the most visually accurate optical flow predictions. These results demonstrate that SLVM not only improves low-light static perception but also effectively enhances motion estimation under poor lighting, showcasing its practical potential.

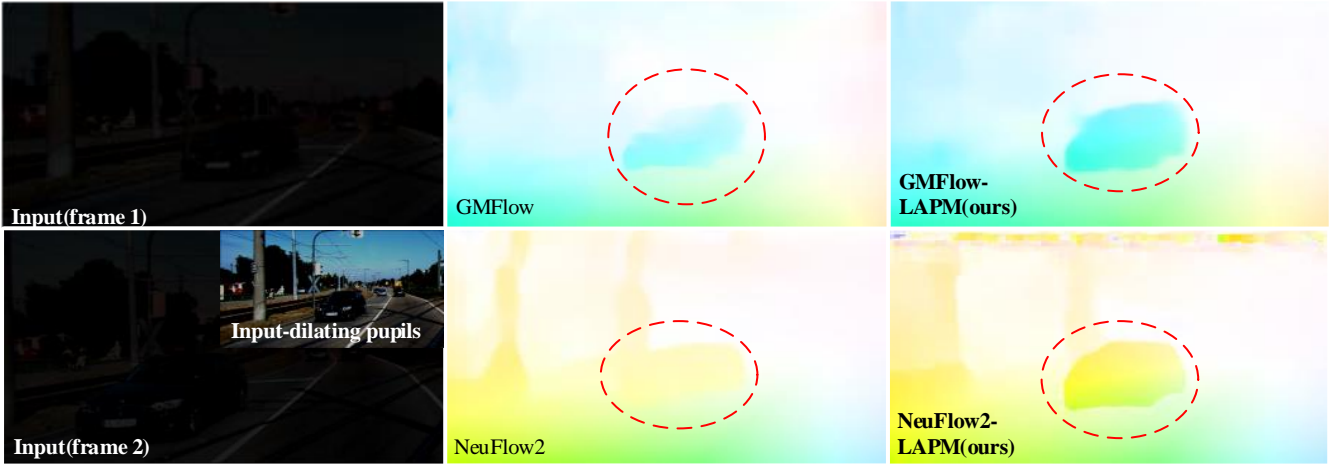

**Figure 10.** Qualitative comparisons on dynamic vision tasks (optical flow estimation). Models originally designed for well-lit conditions generate blurry motion boundaries under low light, whereas SLVM-integrated models exhibit significantly improved optical flow accuracy.

## 5. Discussion

Our study underscores the critical limitations of existing perception systems in low-light traffic scenarios, particularly in terms of generalization, robustness, and efficiency. Through the construction of the Dark-traffic dataset and extensive experiments, we demonstrate the necessity of task-specific low-light benchmarks for meaningful evaluation. Existing public datasets either lack real-world traffic semantics or fail to provide dense instance-level annotations and motion cues, thereby limiting their utility in perception research. The proposed SLVM framework introduces a biologically inspired architecture that effectively addresses visual degradation and the scarcity of reliable cues under low illumination. In contrast to enhancement-based approaches that prioritize pixel-level restoration, SLVM focuses on representational fidelity and task-relevant perception. It integrates luminance-adaptive feature capture, feature-level decomposition, and task-specific separable processing to promote both efficiency and accuracy.

Notably, SLVM demonstrates strong adaptability across static detection, static instance segmentation, and motion-aware flow estimation, validating the benefits of modular and interpretable architectures aligned with biological vision systems. However, SLVM has not yet extended static object perception into the broader domain of dynamic perception in the real physical world—where true motion cues, including unknown object movements and global dynamics, must be perceived and integrated. Pseudo-dynamic perception obtained by stitching static frames into video streams lacks authentic motion information and remains dependent on category-specific recognition. In contrast, biological vision systems inherently integrate object and motion perception in a unified and continuous manner. Moreover, while SLVM is designed with real-time deployment in mind and achieves low latency, there remain trade-offs in practical applications. For instance, the feature decomposition strategy may introduce minor latency overhead when scaling to large, high-resolution input, and the modular architecture requires careful scheduling across parallel hardware to fully realize its efficiency gains. These factors should be considered when deploying on latency-sensitive or resource-constrained platforms such as edge devices. These limitations highlight key directions for future work.

## 6. Conclusion

In this paper, we address the challenge of object perception in low-light traffic environments by proposing a novel biologically inspired framework, SLVM, and introducing the Dark-Traffic dataset—the largest to date with dense annotations for detection, instance segmentation, and optical flow under realistic low-light conditions. Our contributions are threefold: (1) a physically grounded degradation model for constructing realistic low-light data, (2) a task-agnostic but modular architecture informed by biological vision principles, and (3) extensive benchmarking showing that SLVM outperforms state-of-the-art real-time models in both accuracy and efficiency across multiple tasks and datasets.

Our work offers new insights into robust perception under adverse lighting and provides the community with valuable resources for advancing low-light scene understanding. All code and data will be made publicly available to facilitate further research.

**Acknowledgments**

This work is supported by National Natural Science Foundation of China (Grant No. 52172381), Graduate Research Innovation Foundation of Chongqing Jiaotong University (Grant No. CYB240259). The author would also like to thank Mr. Chuanlin Liu, Mr. Penghao Zhu, and Ms. Xinxin Liu (Intelligent Transportation Big Data Center of Chongqing Jiaotong University) for providing GPUs.

# Supplementary Materials

**Appendix A. Training hyperparameters and implementation details of the SLVM**

All experiments were conducted on a workstation equipped with an NVIDIA Tesla A40 GPU and an Intel® Xeon® Gold 6338 CPU @ 2.00GHz × 24, running Ubuntu 20.04. More implementation details and training configurations can be found at https://github.com/alanli1997/slvm.

**Table a**. Training hyperparameters of SLVM on static perception tasks (object detection, instance segmentation)

| hyperparameters | value | hyperparameters | value | hyperparameters | value |
|---|---|---|---|---|---|
| epochs | 200 | optimizer | SGD | warmup epochs | 3 |
| batch size | 16 | learning rate(lr) | 0.01-0.0001 | warmup momentum | 0.8 |
| resolution | 640×640 | momentum | 0.937 | warmup bias lr | 0.1 |
| num_workers | 16 | weight decay | 0.0005 | early stop epochs | 50 |

**Table b**. Training hyperparameters of SLVM on dynamic perception tasks (optical flow)

| hyperparameters | value | hyperparameters | value | hyperparameters | value |
|---|---|---|---|---|---|
| training steps | 10000 | optimizer | AdamW | eps | $1e^{-8}$ |
| batch size | 8 | learning rate(lr) | 0.0002 | amsgrad | False |
| resolution | 320×1152 | betas | (0.9, 0.999) | maximize | False |
| num_workers | 8 | weight decay | 0.0001 | foreach | False |

Detailed backbone structure of the SLVM is reported in Tabel c and Tabel d. The neck and head structure of the SLVM at https://github.com/alanli1997/slvm.

**Table c**. Structure of the SLVM-S. The ConvBlock and SPPF block are from the YOLO-Utralytics and could be replaced by others Convolutional block or Transformer block (e. g. ResBlock or Self-attention block) according the vision task.

| Backbone of the FSL branch | | Backbone of the LAPM branch | |
|---|---|---|---|
| **Block** | **channels, kernel size, stride** | **Block** | **channels, kernel size, stride** |
| Convolution | 16, 3, 2 | LAPM | 1, 2, 2 |
| FSLConv | 32, 3, 2 | LAPM | 1, 2, 2 |
| ConvBlock | 32, 1, 1 | - | - |
| FSLConv | 64, 3, 2 | LAPM | 1, 2, 2 |
| ConvBlock×2 | 64, 1, 1 | - | - |
| FSLConv | 128, 3, 2 | LAPM | 1, 2, 2 |
| ConvBlock×2 | 128, 1, 1 | - | - |
| FSLConv | 256, 3, 2 | LAPM | 1, 2, 2 |
| ConvBlock | 256, 1, 1 | - | - |
| SPPF | 256, [5, 9, 13], 1 | - | - |

**Table d**. Structure of the SLVM-L. The ConvBlock and SPPF block are from the YOLO-Utralytics and could be replaced by others Convolutional block or Transformer block (e. g. ResBlock or Self-attention block) according the vision task.

| Backbone of the FSL branch | | Backbone of the LAPM branch | |
|---|---|---|---|
| **Block** | **channels, kernel size, stride** | **Block** | **channels, kernel size, stride** |
| Convolution | 64, 3, 2 | LAPM | 1, 2, 2 |
| FSLConv | 128, 3, 2 | LAPM | 1, 2, 2 |
| ConvBlock×3 | 128, 1, 1 | - | - |
| FSLConv | 256, 3, 2 | LAPM | 1, 2, 2 |
| ConvBlock×6 | 256, 1, 1 | - | - |
| FSLConv | 512, 3, 2 | LAPM | 1, 2, 2 |

| ConvBlock×6 | 512, 1, 1 | - | - |
| FSLConv | 1024, 3, 2 | LAPM | 1, 2, 2 |
| ConvBlock×3 | 1024, 1, 1 | - | - |
| SPPF | 1024, [5, 9, 13], 1 | - | - |

**Appendix B. Detailed implementation of the LAPM in SLVM**

Given an input RGB image tensor $X \in \mathbb{R}^{H \times W \times 3}$, where $H$ and $W$ denote the image height and width, respectively. The LAPM performs the following steps to simulate pupil dilation and extract light-adaptive texture features.

(I) Pupil dilation-based luminance compensation

We simulate pupil dilation via a pixel-wise amplification of the RGB image channels:

$$\hat{X}_{i,j,c} = \lambda \cdot X_{i,j,c}, \forall c \in \{R, G, B\} \quad (a)$$

where $\hat{X}_{i,j,c}$ is the luminance-compensated pixel value in the enhanced image; $X_{i,j,c}$ is the intensity value of the $c$-th color channel (R: red, G: green, B: blue) at pixel location $(i, j)$; $\lambda$ is the pupil dilation amplification factor (typically $\lambda \in [5.0, 12.0]$), simulating the intensity boost due to biological pupil expansion under low-light conditions.

(II). Grayscale conversion

The enhanced image is then converted to grayscale to preserve luminance structure and reduce chromatic noise. This follows the ITU-R BT.601 standard:

$$X_{i,j}^{gray} = 0.299 \cdot \hat{X}_{i,j,R} + 0.587 \cdot \hat{X}_{i,j,G} + 0.114 \cdot \hat{X}_{i,j,B} \quad (b)$$

This produces a single-channel grayscale map $X_{i,j}^{gray} \in \mathbb{R}^{H \times W}$.

(III). Binary photosensitive mask generation

To simulate photoreceptor thresholding, we apply binary binarization:

$$X_{i,j}^{binary} = \begin{cases} 1, & \text{if } X_{i,j}^{gray} > \tau \\ 0, & \text{otherwise} \end{cases} \quad (c)$$

where $\tau$ is a fixed luminance threshold (e.g., $\tau = 0.02$). The binary mask $X_{i,j}^{binary}$ highlights the light-sensitive areas.

(IV). Texture feature extraction via gated normalization

The binary mask $X_{i,j}^{binary}$ is convolved using a lightweight kernel and passed through gated normalization:

$$X_{i,j}^{texture} = \frac{\gamma \cdot (W * X_{i,j}^{binary})_{i,j} + \beta}{1 + \exp(-(W * X_{i,j}^{binary})_{i,j}) + \epsilon} \quad (d)$$

where $W$ is a $1 \times 1$ learnable convolution kernel; $\gamma$ and $\beta$ are learnable batch normalization scaling and shifting parameters; $\epsilon = 10^{-8}$ ensures numerical stability.

The sigmoid-based denominator softly gates the activation magnitude, ensuring smooth response to local luminance variations.

The output $X_{i,j}^{texture}$ is a light-compensated texture feature map emphasizing informative regions under low illumination and fused back into the main backbone stream for downstream tasks (e.g., detection or segmentation). Notably, the LAPM contains only **4 learnable parameters** and introduces merely **0.002184 GFLOPs** per 640×640 image, making it highly efficient and suitable for real-time deployment.

**Appendix C. The training logs of SLVM by TensorBoard**

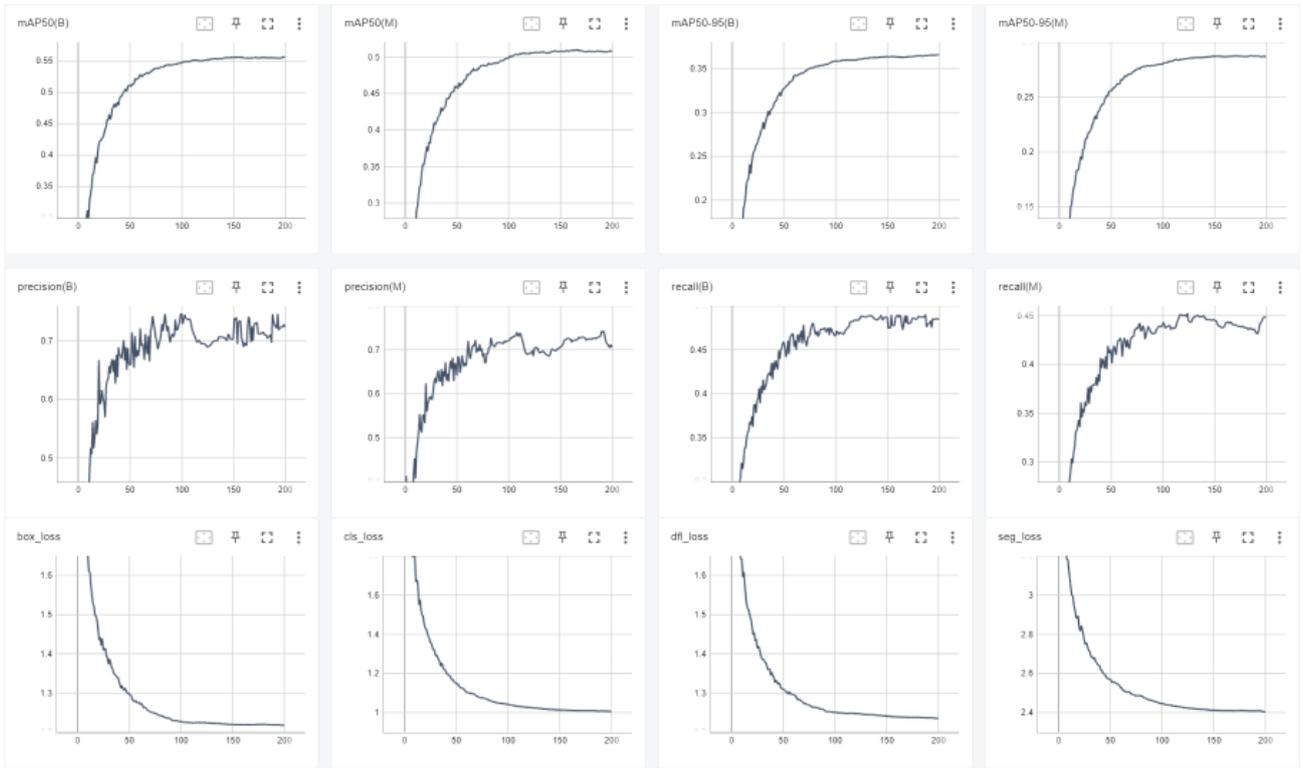

**Figure a.** Visualized training log of the SLVM-S on Dark-traffic (instance segmentation).

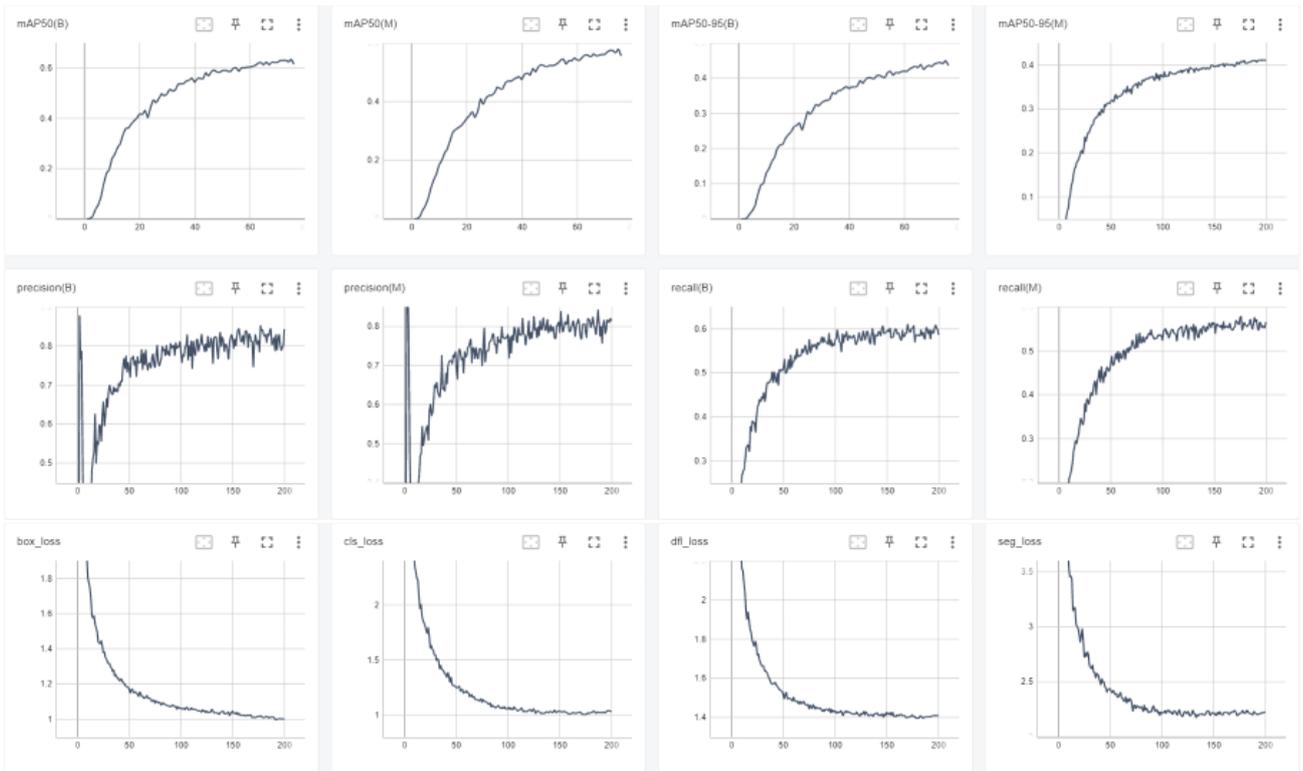

**Figure b.** Visualized training log of the SLVM-L on LIS (instance segmentation).

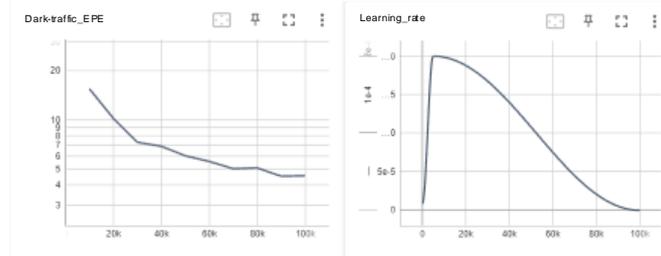

**Figure c.** Visualized training log of the GMFlow-LAPM on Dark-traffic (optical flow). The NeuFlow2 project did not utilize TensorBoard for logging the training process; therefore, the figures do not include training

records for NeuFlow2+LAPM.

In Figures a, b, and c, we provide the visualized training logs of SLVM-S, SLVM-L, and GMFlow-LAPM, respectively, for quick reference. Figures a and b illustrate the curves of prediction metrics ($AP^{box}$, $AP^{mask}$, Precision, and Recall) on the *Dark-traffic* instance segmentation training set, along with the validation losses (Bounding box loss, Segmentation mask loss, Classification loss, and DFL loss). Figure c shows the End-Point Error (EPE) and learning rate curves during training on the *Dark-traffic* optical flow validation set.

Please note that these plots are screenshots of TensorBoard training logs. All checkpoints and reproducible code for SLVM can be found at: https://github.com/alanli1997/slvm